\documentclass[runningheads]{llncs}

\usepackage [USenglish]{babel}
\usepackage [autostyle, english = american]{csquotes}
\MakeOuterQuote{"}
\usepackage{graphicx}
\usepackage{xcolor}
\usepackage[pdftex, colorlinks=true, hyperfootnotes=true, hyperindex=true,
plainpages=false, pagebackref=false, pdfpagelabels=true, pdfstartview=FitH,
linkcolor=blue, citecolor=blue, urlcolor=blue,
bookmarks, bookmarksopen, bookmarksdepth=3]{hyperref}

\usepackage{nth}
\usepackage{amsfonts}

\usepackage{cite}

\usepackage{multirow}

\usepackage{caption}
\usepackage{graphicx}
\usepackage{subcaption}
\usepackage{wrapfig}

\usepackage{calrsfs}
\DeclareMathAlphabet{\pazocal}{OMS}{zplmr}{m}{n}

\usepackage{url}
\usepackage{hyperref}
\usepackage{enumitem}

\usepackage{tabularx}
\newcolumntype{Z}{>{\centering\let\newline\\\arraybackslash\hspace{0pt}}X}
\newcolumntype{Y}{>{\centering\arraybackslash}X}

\usepackage[linesnumbered,ruled]{algorithm2e}

\urldef{\mailsa}\path|{lars.ackermann, julian.neuberger, stefan.jablonski}@uni-bayreuth.de |

\newcommand{\keywords}[1]{\par\addvspace\baselineskip\noindent\keywordname\enspace\ignorespaces#1}

\title{Beyond Rule-based Named Entity Recognition and Relation Extraction for Process Model Generation from Natural Language Text}
\titlerunning{Beyond Rule-based Process Model Generation}
\author{Julian Neuberger, Lars Ackermann\orcidID{0000-0002-6785-8998}, \and Stefan Jablonski}
\authorrunning{Neuberger et al.}
\institute{University of Bayreuth, Germany\\
\email{\{firstname.surname\}@uni-bayreuth.de}\\
}

\begin{document}
\maketitle

\renewcommand{\figurename}{Fig.}
\renewcommand{\figureautorefname}{Fig.}
\renewcommand{\tablename}{Tab.}
\renewcommand{\tableautorefname}{Tab.}
\renewcommand{\sectionautorefname}{Sec.}
\renewcommand{\subsectionautorefname}{Sec.}
\renewcommand{\subsubsectionautorefname}{Sec.}

\begin{abstract}

Process-aware information systems offer extensive advantages to companies, facilitating planning, operations, and optimization of day-to-day business activities. 
However, the time-consuming but required step of designing formal business process models often hampers the potential of these systems. 
To overcome this challenge, automated generation of business process models from natural language text has emerged as a promising approach to expedite this step.
Generally two crucial subtasks have to be solved: extracting process-relevant information from natural language and creating the actual model.
Approaches towards the first subtask are rule based methods, highly optimized for specific domains, but hard to adapt to related applications.
To solve this issue, we present an extension to an existing pipeline, to make it entirely data driven.
We demonstrate the competitiveness of our improved pipeline, which not only eliminates the substantial overhead associated with feature engineering and rule definition, but also enables adaptation to different datasets, entity and relation types, and new domains. 
Additionally, the largest available dataset (PET) for the first subtask, contains no information about linguistic references between mentions of entities in the process description. 
Yet, the resolution of these mentions into a single visual element is essential for high quality process models.
We propose an extension to the PET dataset that incorporates information about linguistic references and a corresponding method for resolving them.
Finally, we provide a detailed analysis of the inherent challenges in the dataset at hand. 
\keywords{Process-aware Information Systems, Process Extraction, Named Entity Recognition, Relation Extraction, Co-Reference Resolution}
\end{abstract}

\section{Introduction}%
\label{sec:introduction}
Automated generation of formal business process models from natural language process descriptions has become increasingly popular~\cite{friedrich2011process, quishpi2020extracting, bellan2023pet, van2018challenges, van2019extracting}. 
This is motivated, for instance, with the comparatively high time expenditure for manually designing said process models. 
Up to 60\% of the total duration in process management projects is spent on the design of process models~\cite{friedrich2011process}. 
Techniques for automated process model generation from natural language text aim to reduce this effort, but have to solve several sub-tasks for this, categorized into two distinct phases:
\textit{(i) The information extraction phase} and \textit{(ii) the process model generation phase}.
During the information extraction phase, techniques recognize process elements (e.g., activities, actors, data objects), extract relations (e.g., sequence-flow relations between activities), and resolve references (e.g., mentions of the same data object).
Building on this information, the process model generation phase creates a concrete process model~\cite{friedrich2011process, quishpi2020extracting, lee2017end}.
The current state of the art for the information extracting phase exhibits two core issues, which we will briefly discuss in the following.

\textbf{Core Issue 1} Existing approaches are largely rule-based, i.e. approaches use manually crafted rules rooted in domain knowledge~\cite{sanchez2021unleashing,van2019extracting,friedrich2011process}.
Rule-based systems usually show remarkable precision and recall for the datasets they are created for. 
However, they \emph{a)} require significant amounts of labor to capture linguistic subtleties, \emph{b)} require deep technical knowledge, as well as knowledge of the target domain, and \emph{c)} are hard to adapt to even minor changes in the underlying data, which leads to unacceptable expansion in the number of required rules~\cite{waltl2018rule}.
Using machine learning, these drawbacks can be resolved, especially deep learning methods have been shown to greatly reduce the amount of effort and domain knowledge required~\cite{li2020survey}.
However, deep learning methods usually need considerable amounts of data for stable training\cite{li2018deep}, something the  field of business process modeling research currently can not provide~\cite{kappel2021leveraging}.
Using less expressive machine learning models constitute a middle ground to this dilemma, as they can be trained stably with orders of magnitude less data.

\begin{figure}[bth!]
    \centering
    \includegraphics[width=\linewidth]{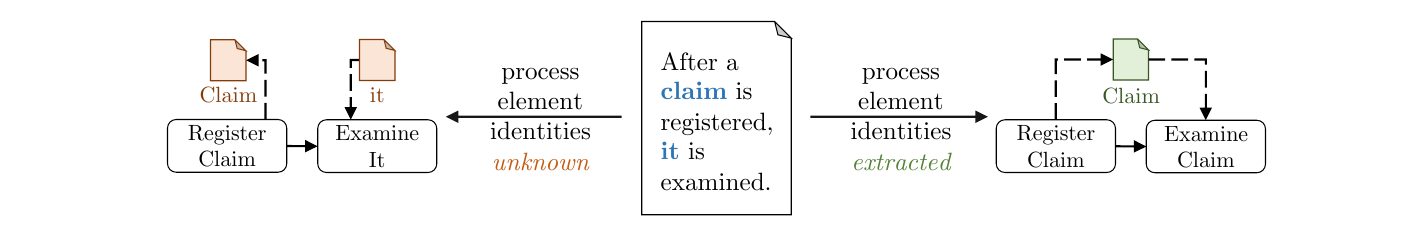}
    \caption{Example for differences between information extraction phase with and without resolving process element identities. Resolving process element identity from their mentions (right) allows generation of correct data flow, without (left) data flow is disjointed.}
    \label{fig:pipeline-differences}
\end{figure}

\textbf{Core Issue 2} Existing approaches are scoped too narrowly~\cite{Qian2020}. 
This includes systems, that do not capture enough information for the generation of complete process models, as well as systems that impose unrealistic assumptions concerning the structure of input text.
Most notably, the currently largest dataset for the information extraction phase (PET~\cite{bellan2023pet}) does not include information about linugistic references between mentions of process elements.
For high-quality process models, resolving references between mentions of the same process element is crucial.
Consider, for instance, the example depicted in \ref{fig:pipeline-differences}. 
For a human reader it is obvious, that both "\emph{a claim}" and "\emph{it}" refer to the same instance of a \emph{claim}.
To automatically extract a process model encoding this knowledge the system needs to resolve the two mentions "\emph{a claim}" and "\emph{it}" to a single entity. 
Without this step, at least two problems manifest in the extracted process models: \textit{(i)} Two distinct data objects for \emph{claim} would be created and, thus, the model is not able to correctly express that both the registration and the examination activities process the same data, and \textit{(ii)} one of the created data objects is labeled \emph{it}, because it is unknown that \emph{it} is a reference to \emph{a claim}. 
Though the \emph{claim} example is solely focusing on the data perspective, entity resolution is also necessary for organizational process elements like, for instance, actors. 
Here, it is necessary to be able to create process models that contain a single actor type for the two mentions of the \emph{claim officer}, which is expressed as a single swimlane in BPMN, for instance. 
In summary, it can be said that entity resolution is what makes it possible in the first place to correctly express relations to data and to actors. 
Following from these two core issues we state three main research questions.

\begin{enumerate}[itemindent=.7cm, label=\textbf{RQ\arabic*}]
    \item\label{RQ1} For comparatively small datasets, such as PET, can machine learning models compete with rule-based methods in terms of precision and recall?
    \item\label{RQ2} Can a pre-trained co-reference resolution approach outperform na\"ive word matching, and can therefore be used as a baseline for resolving linguistic references between process element mentions?
    \item\label{RQ3} Are deep learning methods able to extract process information with precision and recall comparable to rule-based methods and less expressive machine learning models given the same dataset?
\end{enumerate}

Our work proposes an improved pipeline which tackles both of these issues, which we describe in detail in Section~\ref{sec:approach-implementation}.
We propose a relation extraction approach based on established machine learning methods.
Additionally, we extend PET with information about the identity of process element mentions, and provide a baseline approach for resolving process element identities from process element mentions.
We compare our pipeline to the current state of the art of information extraction on PET and show that we outperform it in five out of six relation types, with an absolute increase of 6\% in $F_1$ scores.

The remainder of this paper is structured as follows: In Section~\ref{sec:problem-definition} we formalize the task of process model extraction. In Section~\ref{sec:related-work} we discuss differences to work related to this paper. Our thorough investigation of the PET dataset and the extraction approaches in Section~\ref{sec:results} is based on a rigorous experiment setup introduced in Section~\ref{sec:exp-setup}. 
Short summaries of the answers to our research questions are provided in Section~\ref{sec:conclusion}. 
Both the source code for our experiments and the extended dataset are publicly available\footnote{In case of acceptance, it will be published on GitHub, until then \href{https://drive.google.com/drive/folders/1s-eojDjkyGU1GZpCT1zpK2lc0q6PgPCp?usp=sharing}{it is available here}}, therefore laying the foundation for further focused research.

\section{Task description}\label{sec:problem-definition}
Natural language processing (\emph{NLP}) is a discipline that aims to exploit natural language input data and spans a wide variety of subfields. 
One of these subfields is Information Extraction from human-readable texts. 
In the following, we describe the extraction of process elements and of relations between them as instances of three sub-problems of information extraction, which are \textit{Named Entity Recognition} (NER), \textit{Relation Extraction} (RE), and \textit{Entity Resolution} (ER). 
We then detail the three subproblems with respect to the extended PET dataset as described in Section~\ref{sec:dataset-extension}. 
Each task assumes that the input text has already been pre-processed, i.e. \textit{tokenized}. 



\textbf{Named Entity Recognition} is the task of extracting spans of tokens corresponding to exactly one element from a set of entities~\cite{li2020survey}. 
While NER traditionally only considered extraction of proper nouns, the definition of a named entity now depends on the domain~\cite{sharnagat2014named}.
For the process domain named entities are process relevant facts, such as actors (e.g. \emph{the CEO} vs. \emph{Max}) or activities, e.g. \emph{approve} vs. \emph{the approval}). 
The PET dataset defines a set of seven process relevant facts, aimed at providing a general schema for the task of process model generation from natural language text~\cite{bellan2022guidelines}.
Formally the NER task is extracting a set of triples $M$ from a given list of tokens $T$, so that for each triple $m = (i_s, i_e, t_e) \in M$, the indices $i_s$ and $i_e$ denote start and end tokens of the span in $T$ respectively, and $t_e$ refers to the entity type.
Throughout this paper, we will refer to the triple $m$ a \emph{mention} of an \emph{entity}.
An extracted mention is considered correct, iff its triple has an exact match in the list of ground truth triples given by the dataset.

\textbf{Entity Resolution} extracts a set of unique entities from a given set of mentions $M$. 
This step can be seen as resolving references between mentions of the same process element, which is crucial information for generating useful business process models further down-stream, as shown in Figure~\ref{fig:pipeline-differences}.
Formally the ER task is defined as finding a set of non-empty \emph{mention clusters} $E$, so that each mention $m \in M$ is assigned to exactly one cluster $e \in E$.
These clusters are called \emph{entities}.
To disambiguate between the use of entity as in NER, and entity as used in ER, we will call the result of NER \emph{mentions} from now on, and the result of ER \emph{entities}.
An entity prediction is considered correct, iff the set of contained mentions is exactly the same as the ground truth defined by the dataset.
Entity resolution itself is a super-set of the tasks \emph{Anaphora Resolution}, i.e., back-referencing pronouns, \emph{Coreference Resolution}, and \emph{Cataphora Resolution}, i.e., forward-referencing pronouns~\cite{sukthanker2020anaphora}.
While there are subtle differences and overlap between these sub-fields, this work focuses on coreference resolution.
While the addition of cataphora and anaphora resolution is potentially useful, it does not warrant the additional complexity for our planned baseline, and is therefore out of scope.
Thus, we refer to coreference resolution, whenever we mention the ER task in later sections.
The PET dataset only contains two entity types, where entity identity is relevant: \emph{Actors}, describing a natural person, department, organization, or artificial agent, and \emph{Activity Data}, which are objects or data used by an~\emph{Activity}~\cite{bellan2022guidelines}.
Further details can be found in section~\ref{sec:dataset-extension}.

\textbf{Relation Extraction} is the task of identifying a set of semantic relations $R$ between pairs of entities. 
Current literature distinguishes between global and mention level RE~\cite{pawar2017relation}.
Global RE is the task of extracting a list of entity pairs forming a certain relation from a text, without any additional information.
On the other hand, mention level RE methods are given a pair of entity mentions and the sentence containing them, and have to predict the relation between the two.
The PET dataset contains relation information on mention level, which allows our approach to learn on local level.
There are six relation types defined in the PET dataset, such as \emph{Flow}, which captures the execution order between behavioural elements~\cite{bellan2022guidelines}.
Each relation is formally defined by a triple $r=(m_h, m_t, t_r)$, where $m_h$ is the head entity mention or source of the relation, $m_t$ the tail entity mention or target, and $t_r$ the type of the semantic relation.
This definition implies relations are directed, that is $(m_h, m_t, t_r) \ne (m_t, m_h, t_r)$ for $m_h \ne m_t$.
A predicted relation tuple $r \in R$ is considered correct, iff its triple has an exact match in the list of ground truth triples given by the dataset.

\section{Related Work}~\label{sec:related-work}

This paper is founded on the work presented in \cite{bellan2023pet} and is therefore closely related, as we use, extend, and analyze the data. 
Additionally we adopt the approach to NER, and compare our proposed RE approach to their pipeline. 
They are missing a more in-depth description of their data, especially regarding qualities important for prediction performance, including but not limited to: correlation between a relation's type and its argument types, or the linguistic variability of their data. 
Furthermore, the implementation of their pipeline is not publicly available, impeding further research and development.

There are several approaches related to the baselines we present and analyze in this work. 
An annotation approach based on rule-based pattern matching across the dependency tree representation of a textual process description is presented in \cite{sanchez2021unleashing}, which is then used to generate an event log. 
This allows the extraction of a formal process model via established process mining techniques. 
While it achieves state-of-the-art results, it uses a tagging schema different from the one used in PET, which makes it unfeasible for use in a direct comparison. 
\cite{friedrich2011process} presents a pipeline able to extract formal process models in Business Process Model and Notation (BPMN), and therefore is locked into this process notation language. 
The same limitation holds for the approach presented in \cite{van2019extracting}, which extracts process models utilizing the Declare language. 
PET follows a different tagging scheme and, thus, a direct comparison is not possible. 
In \cite{Qian2020} a neural method for entity and relation classification is proposed, but assumes that relevant text fragments are already extracted. 
This is a significantly easier task, since separating relevant process information from redundant, superfluous, and incidental information, appearing in natural language, is a hard task in itself.
\cite{ackermann2021data} presents an efficient deep learning method using formal meaning representations as an intermediary feature. 
Since they only solve NER, we can not compare their approach with our proposals. 

Due to the strong relation between process extraction and the combined NLP task of NER, ER, and RE, there are several approaches potentially able to solve the process extraction task~\cite{cabot2021rebel,giorgi-etal-2022-sequence,eberts-ulges-2021-jerex,sanh2019hmtl}.
\cite{ackermann2023bridging} studies several approaches built for joint NER, ER, and RE on small documents.
Applying them to the BPM domain entails fragmenting the larger documents of PET properly, as well as dealing with long distance relations, which is out of scope for this paper.
However, we chose Jerex~\cite{eberts-ulges-2021-jerex}, since \cite{cabot2021rebel} and \cite{giorgi-etal-2022-sequence} predict mentions as their textual representation (\emph{surface forms}) only, meaning the span of text containing them might be ambiguous, and therefore token indices not resolvable. 
This violates our definition of mentions (Section~\ref{sec:problem-definition}) and hampers the evaluation of the predictions. 
\section{Process Information Extraction Approach}\label{sec:approach-implementation}

\begin{figure}[bth!]
    \centering
    \includegraphics[width=\linewidth]{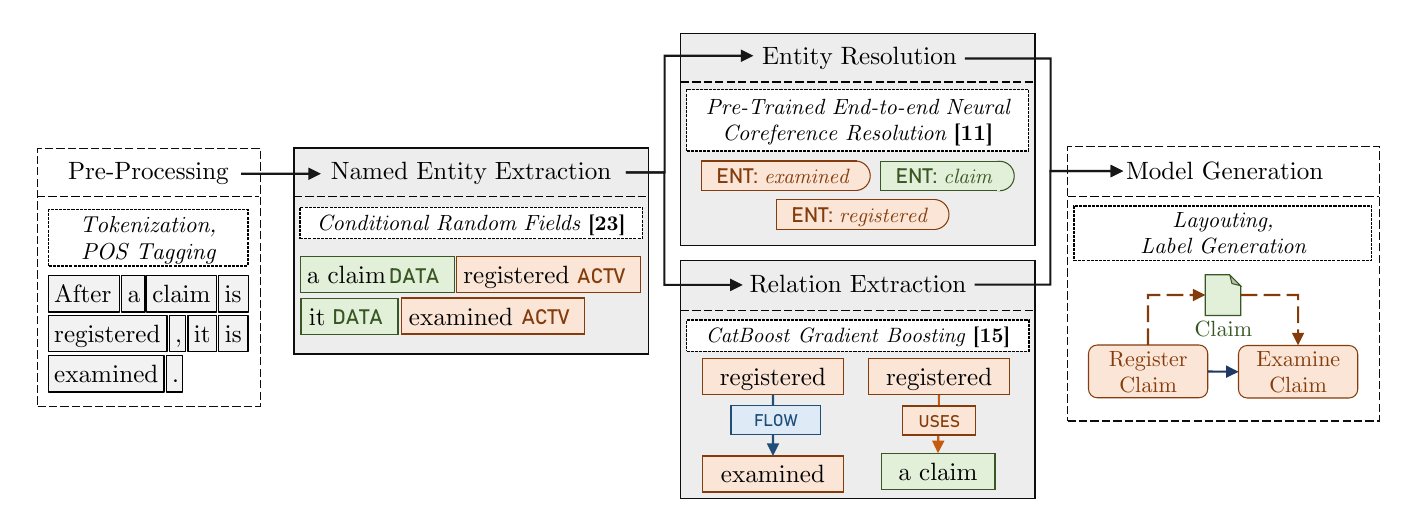}
    \caption{Outline of our proposed extended extraction pipeline.}
    \label{fig:simple-pipeline}
\end{figure}

In the following we present a short overview of the implementation for the three pipeline steps for \emph{NER}, \emph{ER}, and \emph{RE}.
The entire pipeline as we propose it is depicted in Figure~\ref{fig:simple-pipeline}.
We will refer to this pipeline implementation as \emph{Ours} from now on.
We do not detail preprocessing steps, nor the actual synthesis of a business process model, as both are out of scope for this paper.

\textbf{The NER step} is identical to the implementation from \cite{bellan2023pet}.
The approach is based on Conditional Random Fields (\textit{CRF}), a powerful technique for tagging a sequence of observations, here tokens in a text~\cite{wallach2004conditional}. 
Given a sequence of tokens tagged in this way, we then resolve mentions, where each mention contains a set of token indices and the predicted process element type.

\begin{figure}
    \centering
    \includegraphics[width=\textwidth]{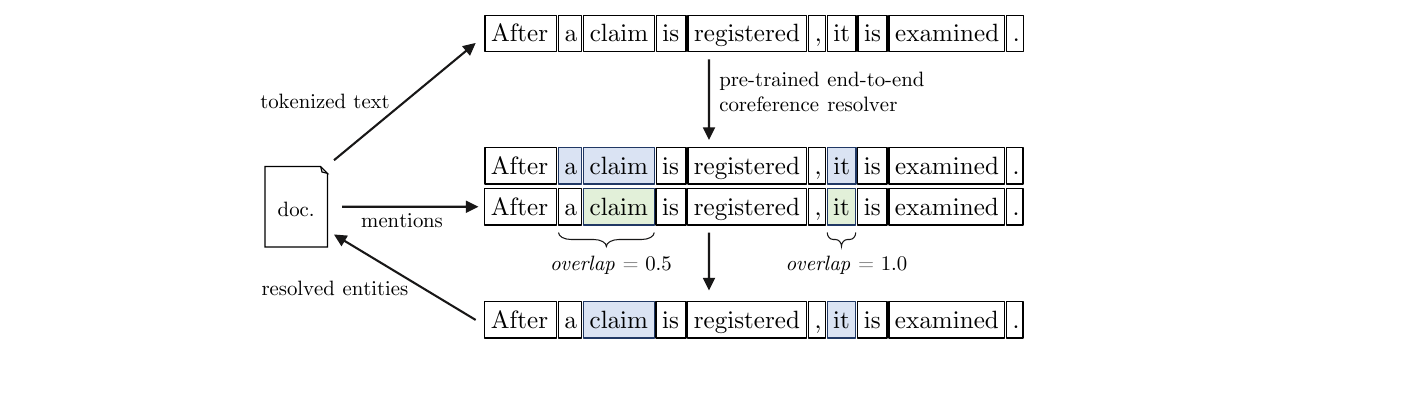}
    \caption{Example for our ER method based on a pretrained end-to-end neural coreference resolver. Predicted coreferent text spans \emph{a claim} and \emph{it} are accepted and resolved to an entity containing the mentions \emph{claim} and \emph{it}, since both text spans overlap at least 50\% with the mention's texts.}
    \label{fig:neural-coref}
\end{figure}

We implemented two modules for \textbf{the ER step}, namely a \emph{naive ER method}, and a method based on \emph{pre-trained end-to-end neural coreference resolution}, as described in~\cite{lee2017end} and implemented in spaCy\footnote{See \url{https://explosion.ai/blog/coref} for more details.}.
The naive ER method, which we will call \emph{naive ER} for short, iteratively selects the best matching mentions with identical NER tags.
The match of two mentions is calculated based on the percentage of \emph{overlapping}, i.e., the fraction of shared tokens over the total number of tokens.
Ranking mention pairs by this score, the naive ER method merges mentions into clusters.
If one of the selected mentions already is part of a cluster, the other mention is added to that cluster as well.
If both selected mentions are part of a cluster, the clusters are merged.
This is repeated until there are only matches left, which overlap less than some threshold $o$.
We ran an optimization to select this overlap optimally and chose $o=0.5$
The pre-trained end-to-end neural coreference resolution module, which we will call \emph{neural ER} from now on, predicts co-referent spans of text, i.e. spans of text referring to each other.
It does so without any domain knowledge, i.e. knowledge about mentions of process elements extracted in prior steps. 
We then align these predictions with mentions. 
Here we discard predictions, if \textbf{(1)} the corresponding span of text is not a mention at all, \textbf{(2)} the corresponding span of text does not overlap with a mention's text by a certain percentage $\alpha_m$, \textbf{(3)} the mention corresponding with the predicted span of text was not tagged with the majority tag of other mentions of this entity, or \textbf{(4)} not at least a certain portion $\alpha_c$ of predicted text spans was previously accepted.
We optimize these parameters using a grid search approach, choosing $\alpha_c=0.5$ and $\alpha_m=0.5$.
A simple example of this process is shown in Figure~\ref{fig:neural-coref}

\begin{figure}
    \centering
    \includegraphics[width=.8\linewidth]{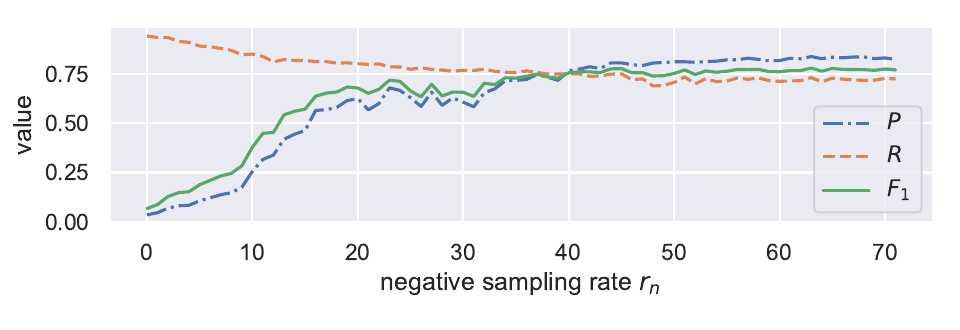}
    \caption{Values of metrics $P$, $R$, and $F_{1}$ for different negative sampling rates $r_n$.}
    \label{fig:p-vs-r-vs-f1}
\end{figure}

Finally the \textbf{RE step} extracts relations between mentions using CatBoost, a gradient boosting technique for classification using numerical, as well as categorical data\cite{prokhorenkova2018catboost}. 
We call this module \emph{BoostRelEx} for short in following sections.
For each combination of head and tail mention of a relation we build features containing tags, distance in tokens and in sentences between them, and a number $c$ of neighboring mention tags as context. This feature set is then presented to the model, which predicts a class for it. 
Classes are the set of relation tags and an additional \textit{nothing} tag to enable the model to predict that there is no relation between two mentions. 
During training we present each of the mention combinations containing a relation to the model exactly once per iteration, as well as a given number of negative examples. 
These negative examples only consist of mention combinations, where corresponding entities do not have a relation. 
This concept, called negative sampling, is important, as there are many more mention combinations without a relation between them (44,708), as there are ones with one (1,916).
Without negative sampling the precision of our relation extraction module would be extremely low, visualized in Figure~\ref{fig:p-vs-r-vs-f1}. 
For each positive sample we select $r_n$ randomly drawn negative ones. 
Increasing $r_n$ has a positive impact on the accuracy with which the model predicts the existence of relations between given pairs of mentions, which is called the \emph{precision} $P$.
Since the model learns it has to reject some mention combinations, it also inevitably rejects correct combinations.
Following directly from this, the model misses more combinations of mentions, where a relation actually would have existed, thus resulting in a lower \emph{recall} $R$.
The harmonic mean between the two scores $R$ and $P$ gives us a good idea of the model's performance.
We discuss this metric in more detail in section~\ref{sec:experiment-setup-evaluation}.
We train the \emph{BoostRelEx} module for $i=1000$ iterations, which is the most computationally intensive step in the whole pipeline, taking about 25 minutes on an Intel i9-9900K CPU @ 3.60GHz, using a negative sampling rate of $r_n=40$ and context size of $c=2$. A sampling rate $r_n \geq 40$ improves the result quality significantly.

\section{Experiment Setup}\label{sec:exp-setup}



In the following we describe the extension of the original PET dataset accompanied with dataset statistics (Section~\ref{sec:dataset-extension}). To enable empirical evaluation Section~\ref{sec:experiment-setup-evaluation} introduces performance measures that are most adequate for the task and the concrete dataset. 

\subsection{Dataset}\label{sec:dataset-extension}

The PET dataset is presented in detail in~\cite{bellan2023pet}, in the following we will only discuss aspects of this dataset directly related to our extension and analysis.
PET contains a total of 45 documents, with seven entity types, and six relation types.
To facilitate the entity resolution task described in Section~\ref{sec:problem-definition}, we assign each mention of a process element to a cluster\footnote{All clusters are defined by two experts, with the help of a third for cases, where their initial annotations differed.}.
This resulted in a total of 163 clusters with two or more mentions, of which there are 75 clusters of \textit{Activity Data} mentions, and 88 clusters of \textit{Actor} mentions.
All other entity types and the remaining \textit{Activity Data} and \textit{Actor} mentions belong to clusters with only a single mention.

We define the \emph{intra-entity distance} as the maximum of each mention's minimal distance to each other mention in the entity. 
This gives us the largest span an extraction method has to reason over, in order to detect two mentions as part of the same entity. 
Averaged over all entities this measure is 31.93 tokens for \textit{Activity Data} elements and 54.84 tokens  for \textit{Actors}. 
Distances between referent mentions are significantly longer for \textit{Actors}, indicating that they possibly are harder to extract. 
Our experiments seem to support this notion, as shown in Figure~\ref{fig:radar-scenario-1-2-3} c) and d), but further analysis may be required to come to a conclusive rationale.

\begin{figure}[hbt]
\centering
\begin{subfigure}[t]{.75\textwidth}
  \centering
  \includegraphics[width=\linewidth]{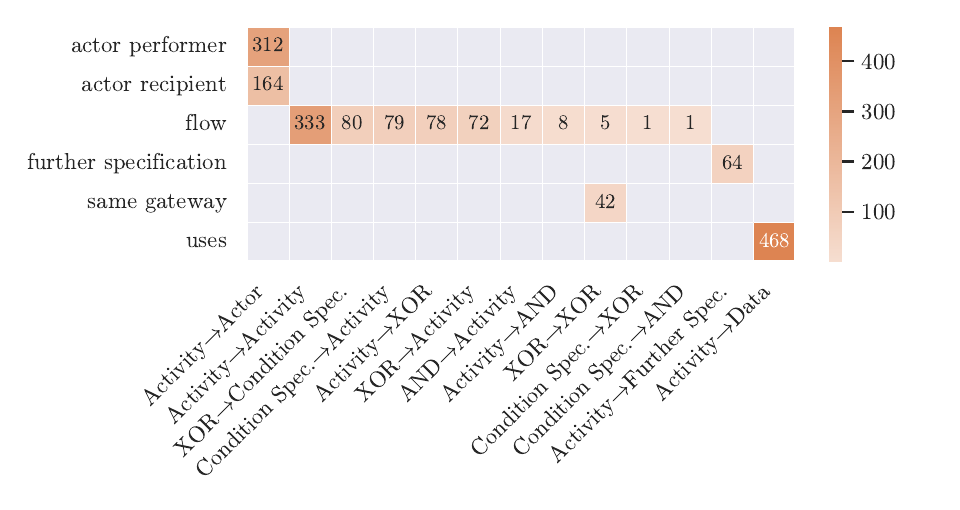}
  \caption{}
  \label{fig:pet-relation-heatmap}
\end{subfigure}%
\hfill%
\begin{subfigure}[t]{.25\textwidth}
    \centering
    \includegraphics[width=\linewidth]{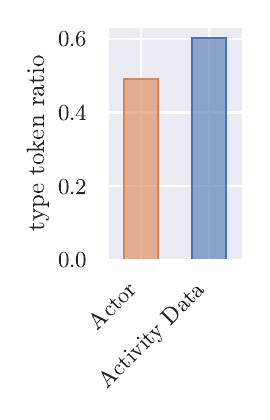}
    \caption{}
    \label{fig:pet-entity-ttr}
\end{subfigure}
\caption{\ref{fig:pet-relation-heatmap} shows the number of relations aggregated by argument types denoted with $\emph{head}\rightarrow\emph{tail}$.
Only combinations where at least one relation exists are shown. \ref{fig:pet-entity-ttr} shows the mean type-token ratio for mention clusters with at least two mentions.}
\end{figure}

%
%

Intuitively, resolving references between mentions of an entity, is easier, when the texts of those mentions are very similar.
Consider, for example, two entities, both made up of two mentions each.
One entity has the mentions \emph{"a claim"} and \emph{"the claim"}, while the other has the mentions \emph{"the claimant"} and \emph{"a applicant"}.
Resolving the first entity should be much easier, since its mentions share common text.
Thus, calculating the lexical diversity of entities of a given type lets us predict how hard it is to extract them without errors.
The \emph{type-token ratio} (TTR) can be used to measure the lexical diversity of a given input text~\cite{richards1987type}.
It is calculated as the ratio between unique tokens and total number of tokens.
High ratios imply very diverse phrases, while low ratios indicate very uniform text.
We select all entities, which contain at least two mentions, and calculate the TTR for each of them.
Take for example the entity consisting of the three mentions \emph{"a claim"}, \emph{"the claim"}, and \emph{"it"}. 
Its TTR would therefore result in $TTR = \frac{4}{5} = 0.8$.
We then calculate the mean of these TTR values, split by entity type.
On average, \textit{Activity Data} mention clusters exhibit higher type-token ratios compared to \textit{Actors}, as visualized in Figure~\ref{fig:pet-entity-ttr}.
This result is leading us to assume \textit{Actors} should be easier to resolve. 
Our experiments support this notion, as can be seen in Figure~\ref{fig:radar-scenario-1-2-3} c).

Figure~\ref{fig:pet-relation-heatmap} shows the distribution of relation types depending on the types of their arguments. 
For relations of type \textit{Actor Performer}, \textit{Actor Recipient}, \textit{Same Gateway}, and \textit{Flows}, knowing the types of their arguments is no discriminating feature. 
For these cases, a data driven approach, such as the one we propose in this paper, is very useful, as complex rules are inferred from data automatically, saving a lot of manual effort.
In contrast, there are also relation types, where their type can directly be inferred from their argument's types, e.g. all relations that have an \textit{Activity} as head argument, and an \textit{Activity Data} element as tail, are of type \textit{Uses}. 
This is hardly surprising when factoring in domain knowledge, as in \textit{PET} \textit{Activity Data} is only used with \textit{Activities}.
Predicting relations of those types is therefore more a matter of detecting them (\emph{recall}), rather than correctly classifying them (\emph{precision}).

\subsection{Compared Approaches}%
\label{sec:experiment-setup-comparison}

We compare our proposed pipeline to the baseline presented in~\cite{bellan2023pet}, extended with our ER module.
The pipeline looks very similar to ours visualized in Figure~\ref{fig:simple-pipeline}, but instead of the \emph{BoostRelEx} module, it uses a rule-based relation extractor, which we will denote \emph{RuleRelEx}.
These rules are defined in~\cite{bellan2023pet}, but have no public implementation, to our knowledge our code is the first executable version available to the community.
There are a total of six rules, which are applied to documents in order. 
This means that rule 1 takes precedence over e.g. rule 3, which relies on this fact, as it needs information about previously extracted \emph{Flow} relations.
We will denote this pipeline with \emph{Bellan + ER} from now on.

Answering \ref{RQ3} requires a deep learning approach, which is able to extract mentions, entities, and relations. 
\emph{Jerex}~\cite{eberts-ulges-2021-jerex} is suitable for this task, as it is a jointly trained end-to-end deep learning approach, and promises to reduce the effect of error propagation.
Jerex takes raw, untokenized text as input, tokenizes it, and produces predictions for mentions, entities, and relations between them.
It is state of the art for the \emph{DocRed} dataset~\cite{yao2019docred}, which is a large benchmark dataset for the extraction of mentions, entities, and relations from documents -- a task description very similar to the one we gave in section~\ref{sec:problem-definition}.
Furthermore, Jerex is able to extract the exact location of mentions inside the input text, unlike competing approaches, which only extract the text of mentions\footnote{See for example the discussion \url{https://github.com/Babelscape/rebel/issues/57}}.
While this drawback may not be as relevant in applications where only the text of a process element is interesting, for the task of business process generation, i.e., the task of generating human-readable, rich labels for activities in a BPMN process model needs the text surrounding a predicted \emph{Acitvity}~\cite{friedrich2011process}.
\subsection{Evaluation}%
\label{sec:experiment-setup-evaluation}

For performance evaluations of existing baselines, as well as our contributions, we adopted the evaluation strategy from \cite{bellan2023pet}. 
This means we run a 5-fold cross validation for the entire pipeline and average individual module scores. 
Errors made by modules during prediction are propagated further down the pipeline, potentially even amplifying in severity, as down-stream modules produce errors themselves as a result. 
To evaluate a given module's performance in isolation, we inject ground-truth data instead of predictions as inputs.
This leads to a total of five different scenarios, for which results are discussed in detail in Section~\ref{sec:results}. 
These scenarios are \textbf{(S1)} \textit{entity resolution} using predictions from the \textit{Mention Extraction} module, and \textbf{(S2)} using ground-truth mentions. 
Furthermore, \textit{relation extraction} \textbf{(S3)} using entities predicted by the pipeline thus far, \textbf{(S4)} using entities predicted during \textit{entity resolution} using ground-truth mentions, and finally \textbf{(S5)} using ground-truth entities.

In each case we use the $F_{1}$ score as a metric, as it reflects the task of finding as many of the expected mentions, entities, and relations as possible (\textit{recall} $R$), without sacrificing \textit{precision $P$} in type or existence prediction. 
$F_{1}$ is then calculated as the harmonic mean of $P$ and $R$, i.e., $F_{1}=\frac{2 \cdot P \cdot R}{P + R}$. 
As there is more than one class within each prediction task, $F_{1}$, $P$, and $R$ have to be aggregated.
Throughout Section~\ref{sec:results} we use the micro averaging strategy, which calculates $P$ and $R$ regardless of a given prediction's class. 
This strategy favours classes with many examples, as high scores in those may overshadow bad scores in classes with few examples.
Should this be of concern, the macro averaging strategy can be used, where $P$, $R$, and $F_{1}$ are calculated for each class separately and averaged afterwards.
We argue that it is most useful to find as many process elements as possible regardless of their type, i.e., it is better to find 90\% of all \textit{Activities} and only 10\% of all \textit{AND Gateways}, instead of 50\% of all elements, as there are 501 \textit{Activities} and only 8 \textit{AND Gateways} in PET~\cite{bellan2023pet}.
As such the micro $F_{1}$ score is better suited to the task.

Following the task description in Section~\ref{sec:problem-definition}, we use the following matching strategies.
We count a \emph{mention} as correctly predicted, iff it contains exactly the same tokens, as the corresponding ground-truth label, and has the same tag. 
We count an \emph{entity} as correctly predicted, iff it contains exactly the same mentions, as the ground-truth label. 
Finally, we count a \emph{relation} as correctly predicted, iff both its arguments, and its tag match the ground-truth label. 
Therefore, e.g., a single missing "\emph{the}" in the mention "\emph{the claim}" would render this mention prediction incorrect, as well as all entities and relations that refer to it.
This effect is called \emph{error propagation} and is the reason why we opted for several scenarios that evaluate modules in isolation, or with some degree of ground-truth input, such as in \textbf{(S4)}. 
It may be, that users are fine with slightly less precise predictions, especially if they only miss inconsequential tokens, such as determiners.
Surveying how users rank the importance of different levels of precision is out of scope of this paper and part of future work.

\section {Results}\label{sec:results}

The following section reports results for the experiments and scenarios defined in the previous Section~\ref{sec:exp-setup}. 
Based on these results, it provides answers for the research questions posed in Section~\ref{sec:introduction}.
In section~\ref{sec:entity-resolution-performance} we provide results for the ER step and compare the naive approach to the one based on pretrained end-to-end neural coreference resolution, both for the modules in isolation (scenario \textbf{(1)}) and based on predictions of the NER module (scenario \textbf{(2)}).
Section~\ref{sec:relation-extraction-performance} presents the results for experiments with the RE step in the end-to-end pipeline setting (scenario \textbf{(3)}), and in isolation (scenario \textbf{(5)}).
Finally, we discuss several factors that affect the quality of RE results in \ref{sec:performance-analysis}, such as the effects of error propagation (scenarios \textbf{(4)} to \textbf{(6)}).

\subsection{Entity Resolution Performance}%
\label{sec:entity-resolution-performance}

We calculate the $F_1$ scores for all mention clusters with at least two mentions, since resolving single mention clusters is trivial.
Figure~\ref{fig:radar-scenario-1-2-3} d) visualizes the difference between the two approaches.
Overall, the naive version reaches $F_{1}=0.26$, while our proposed pretrained method outperforms it significantly and reaches $F_1=0.52$.
This stark difference is rooted in the fact, that we use exact matching, where a single missing or superfluous mention in a cluster renders the entire prediction incorrect.
By design, the naive approach is unable to resolve anaphoras and cataphoras, i.e., back-referencing and forward-referencing pronouns.
This means that every entity containing at least one anaphora, or cataphora, will be predicted incorrectly.
Using the results from the NER step reduces performance greatly, similar as in the RE step.
Based on the results from our experiments we conclude that a naive ER method is not feasible, and significant gains in performance can be achieved by using neural methods.
It would be interesting, if fine-tuning the pretrained model would result in improved accuracy.
Additionally, using information about mentions extracted in the NER step could be integrated into ER, instead of using a task-agnostic model, as we do currently.
These considerations are currently out of scope, as the work on ER in this paper is aimed at bridging the gap between the current state of the art in machine-learning focused data for extracting business process models from natural language text (PET), and the needs of down stream methods.
The discussion in this section leads us to answering \ref{RQ2}: 
The pretrained coreference resolution approach we presented is able to outperform naive text matching significantly, and is a useful baseline for resolving entities from mentions in the setting of business process model generation from natural language text.

\subsection{Relation Extraction Performance}%
\label{sec:relation-extraction-performance}

\begin{figure}[bth!]
\centering
\includegraphics[width=\linewidth]{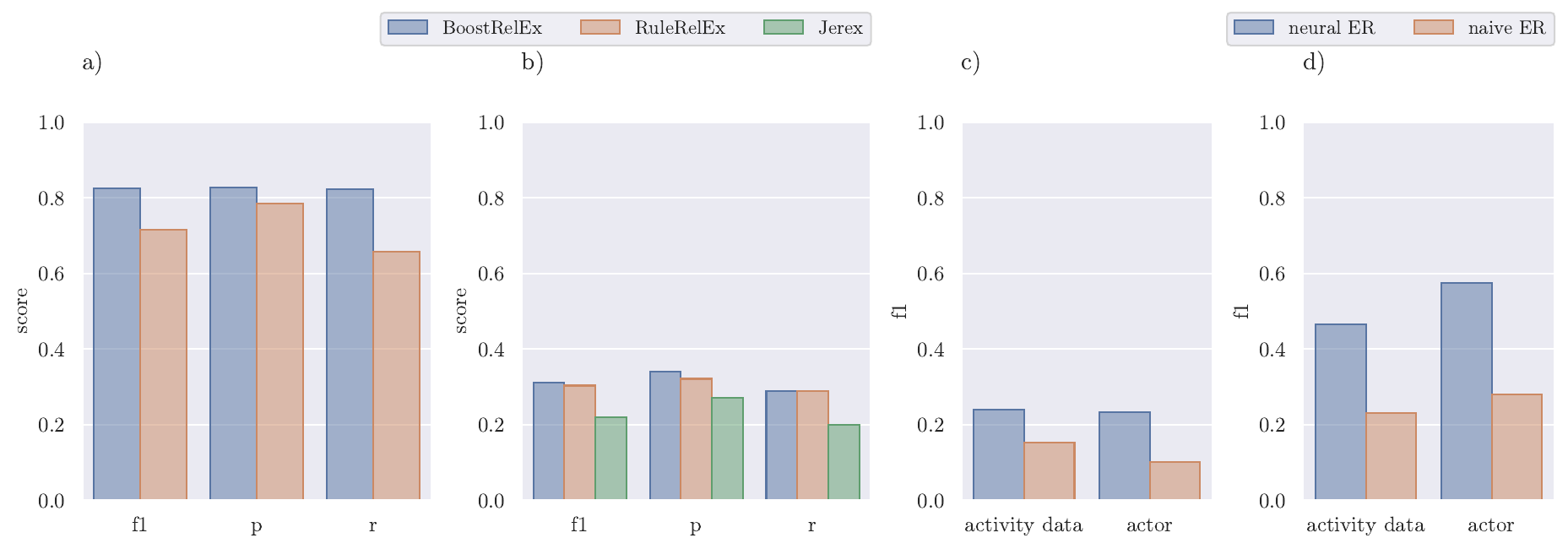}
\caption{a) shows the comparison between \emph{BoostRelEx} and \emph{RuleRelEx}. b) shows the performance of end-to-end runs of our proposed pipeline \emph{Ours}, and \emph{Bellan + ER}. c) compares the performance of the \emph{naive ER} and \emph{neural ER} using the result of the NER step. d) shows the same comparison as c), but based on ground truth mentions.}
\label{fig:radar-scenario-1-2-3}
\end{figure}

Our proposed \emph{BoostRelEx} step clearly beats \emph{RuleRelEx} from~\cite{bellan2023pet} by $F_1=0.10$, $P=0.04$, and $R=0.16$ in our experiments.
This is visualized in Figure~\ref{fig:radar-scenario-1-2-3}, while Table~\ref{tab:overall-scores-isolation} lists exact numbers.
\emph{BoostRelEx} profits greatly from correct predictions during the \emph{NER step}, as is evidenced by greatly reduced performance when running our proposed pipeline end to end, as well as \emph{Bellan + ER}.
While our pipeline is still able to beat \emph{Bellan + ER} in our experiments, the margin is narrowed substantially, with a difference of $F_1=0.01$, $R=0.02$, and equivalent recall.
One reason for this drastic performance loss, is the exact matching strategy we employ. 
A missing, superfluous, or misclassified mention will produce errors during the RE step, as a relation is only considered correct, if all involved mentions are correct (cf. section~\ref{sec:experiment-setup-evaluation}).

Considering the strong effect error propagation has on \emph{BoostRelEx}, using a jointly trained end-to-end model seems natural.
In section~\ref{sec:experiment-setup-comparison} we presented \emph{Jerex} as a promising candidate.
Yet, following from our experiments, \emph{Jerex} is not able to compete, and performs significantly worse, with a difference of $F_1=0.11$, $P=0.14$, and $R=0.02$, compared to our pipeline.
We suspect that this is rooted in PET's small size, as well as the huge number of trainable parameters of Jerex.
We therefore have to answer research question \ref{RQ3} with \emph{No}.

\begin{table}[htb]
    \begin{subtable}[t]{.45\linewidth}
    \centering
    \begin{tabular}{lccc}
    \hline
                                            & $P$               & $R$               & $F_{1}$           \\
    \hline
    Jerex\cite{eberts-ulges-2021-jerex}     & 0.20              & 0.27              & 0.22              \\
    Bellan\cite{bellan2023pet} + ER         & 0.32              & 0.29              & 0.30              \\
    Ours                                    & \textbf{0.34}     & 0.29              & \textbf{0.31}     \\
    \hline
    \end{tabular}
    \caption{}
    \label{tab:overall-scores-complete}
    \end{subtable}%
    \hfill%
    \begin{subtable}[t]{.45\linewidth}
    \centering
    \begin{tabular}{lccc}
    \hline
                                            & $P$               & $R$               & $F_{1}$           \\
    \hline
    Baseline\cite{bellan2023pet} + ER       & 0.79              & 0.66              & 0.72              \\
    Ours                                    & \textbf{0.83}     & \textbf{0.82}     & \textbf{0.82}     \\
    \hline
    \end{tabular}
    \caption{}
    \label{tab:overall-scores-isolation}
    \end{subtable}%
\caption{\ref{tab:overall-scores-complete}: Overall performance for Jerex, the PET baseline, and our proposed enhanced pipeline. \ref{tab:overall-scores-isolation}: Performance of our proposed machine learnt and the rule-based baseline relation extraction modules in isolation.}
\end{table}

\begin{figure}[hbt]
\centering
\includegraphics[width=\linewidth]{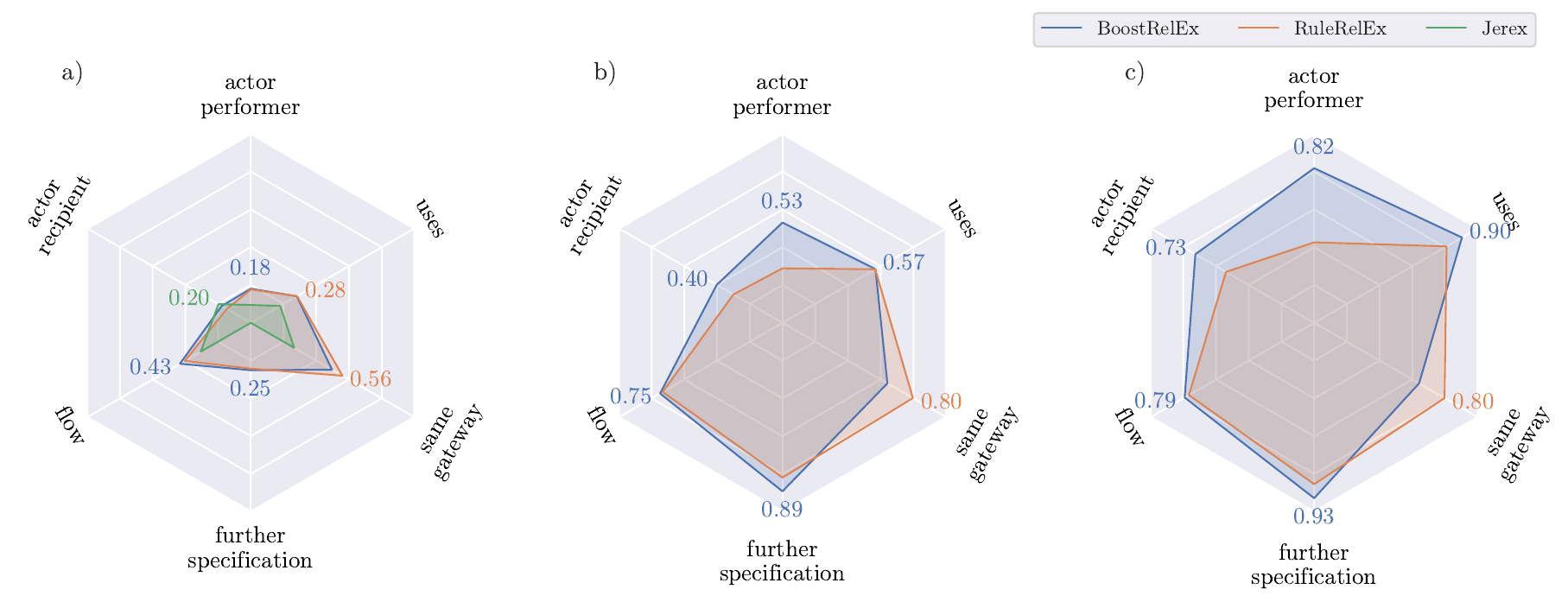}
\caption{a) Results for relation extraction by relation type for scenario \textbf{(4)}, the complete pipeline, b) scenario \textbf{(5)}, relation extraction using entities resolved from perfect mentions, c) scenario \textbf{(6)}, relation extraction from perfect entities.}
\label{fig:radar-scenario-4-5-6}
\end{figure}

Figure~\ref{fig:radar-scenario-4-5-6} breaks down the $F_1$ score by relation type.
Following these results we conclude that the dataset PET is not yet suitable to train deep learning models in a supervised manner.
The amount of data currently available makes stable convergence not possible, preventing the creation of useful models. 
To alleviate the issue of low data, further research into the use of pretrained models, such as LLMs is warranted.
These models make use of large quantities of unlabeled data to learn the structure and makeup of natural language.
They are then either employed in a zero-shot setting (never explicitly trained for the task), few-shot setting (fine-tuned on small quantities of task specific data), or composited into new models (used for extracting useful features from natural language text).
In \cite{DBLP:conf/aiia/BellanDG22} the authors discuss the feasibility of pretrained LLMs and in-context learning for extracting process relevant facts and relations, which shows promise for the use in the business process model generation task in a low data regime.

A significant portion of the improvements we present in this work, come from the better extraction of \emph{Actor Recipient} and \emph{Actor Performer}, as well as the \emph{Uses} relations.
\emph{BoostRelEx} is clearly outperformed by \emph{RuleRelEx} when extracting the \emph{Same Gateway} relation type.
A possible reason for this fact is that \emph{RuleRelEx} uses information about already extracted \emph{Flow} relations (cf. section~\ref{sec:experiment-setup-comparison}), which is not possible for our machine learnt approach, as it extracts all relations at once.
Defining an order of extraction for relation types would defeat the purpose of using our method in the first place: It would be tightly coupled to the dataset and could not be applied easily to others.
The overall performance is not affected very much by this, as there are only a handful of examples for the \emph{Same Gateway} relation.
Still, further research into features useful for properly extracting \emph{Same Gateways} is needed, as well as possible training techniques that allow learning more complex rules. 
Promising features are e.g., synonyms and hypernyms for key phrases of mentions. Furthermore, training the model in multiple passes, each time refining its predictions, could be useful in predicting relations, that feature mutual exclusivity, such as the \textit{Same Gateway} relation does in PET.

\subsection{Performance Analysis}%
\label{sec:performance-analysis}

Gradually reducing the quality of inputs to the \emph{BoostRelEx} and \emph{RuleRelEx} steps results in gradually worse performance, a clear indication of error propagation (cf. section~\ref{sec:experiment-setup-evaluation}).
Using ground-truth mentions from the dataset, but entities predicted by the \emph{neural ER} step, results in a drop in $F_1$ scores of about $0.20$ for \emph{BoostRelEx} and $0.12$ for \emph{RuleRelEx}.
Introducing errors even further down stream, by using the \emph{NER} module, i.e., running the complete pipeline end-to-end results in a drop in $F_1$ of $0.51$ for \emph{BoostRelEx} and $0.42$ for \emph{RuleRelEx}.
Figure~\ref{fig:radar-scenario-4-5-6} visualizes this performance degradation for each relation type individually.
Further studies regarding less strict evaluation is warranted, as described in Section~\ref{sec:experiment-setup-evaluation}, to get a less conservative assessment of prediction quality.

\begin{figure}[hbt]
\centering
\begin{subfigure}[t]{.33\textwidth}
  \centering
  \includegraphics[width=\linewidth]{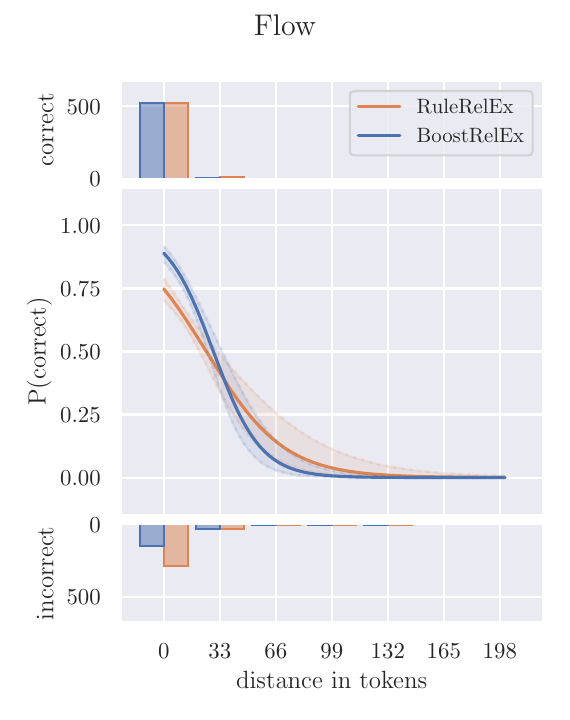}
  \caption{}
  \label{fig:distance-vs-correct-flow}
\end{subfigure}%
\hfill%
\begin{subfigure}[t]{.33\textwidth}
  \centering
  \includegraphics[width=\linewidth]{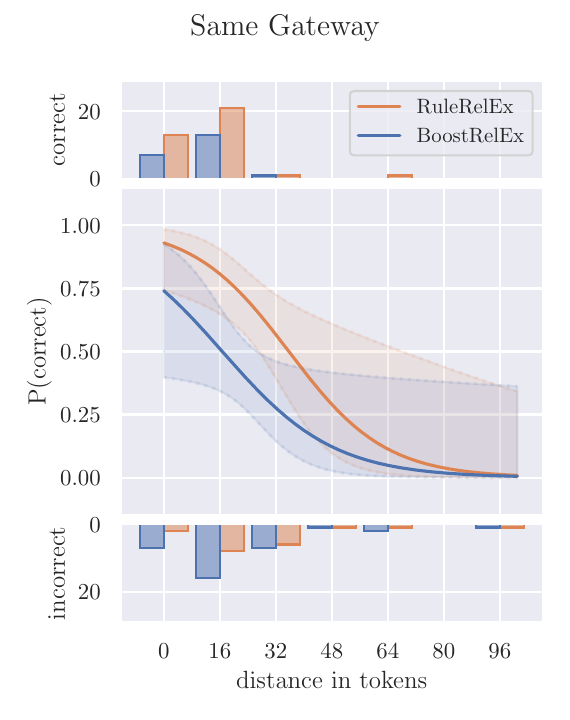}
  \caption{}
  \label{fig:distance-vs-correct-same-gateway}
\end{subfigure}%
\hfill%
\begin{subfigure}[t]{.33\textwidth}
  \centering
  \includegraphics[width=\linewidth]{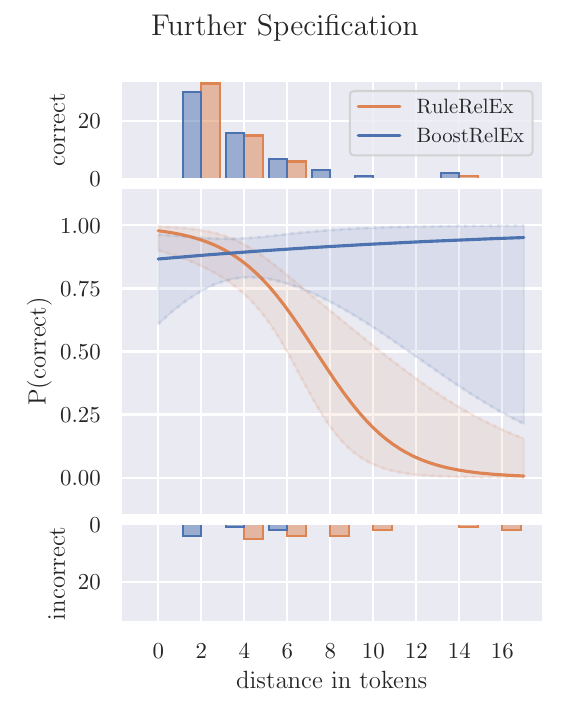}
  \caption{}
  \label{fig:distance-vs-correct-further-specification}
\end{subfigure}
\caption{Logistic regression fits for correlation between correctness of a prediction and the distance in tokens between its arguments. The top and bottom bar plots show the number of correct (top) and incorrect (bottom) predictions. The main plot shows the fitted logistic regression as a solid line, and the 95\% confidence intervals as a transparent channel.}
\label{fig:distance-vs-correct}
\end{figure}

The quality of inputs is not the only factor in relation extraction quality.
We found that the distance between a relation's arguments is also negatively correlated with correctness. 
Longer distance between the head and tail entity of a relation increases the likelihood of misclassifying it, or not detecting it at all.
We calculate the distance of a relations arguments as the minimal distance between the two entity's mentions.
Examples for this effect are shown in Figure~\ref{fig:distance-vs-correct}.
We created datasets from all predictions of each approach, with tuples of the form $(distance, o)$, where $o=1$ denotes a correct prediction, and $o=0$ an incorrect prediction.
We then fitted a logistic regression model to these datasets using the \emph{statsmodels}\footnote{See \url{https://www.statsmodels.org/stable/generated/statsmodels.discrete.discrete_model.Logit.html}.} python package.
A logistic regression model tries to predict an outcome (response variable) via some input variable (predictor variable).
It uses the logistic regression, which is given by $y=\frac{1}{1+e^{-(\beta_0+\beta_{1}x})}$, and chooses $\beta_0$ and $\beta_1$ in such a way, that the model predicts the observed outcome $y=o$ given an input $x$ as best as possible.
We can then use the resulting curve to discuss how well an approach is able to predict certain relation types.

The \emph{Flow} relation can be solved very well for short distances by both \emph{BoostRelEx} and \emph{RuleRelEx}.
A very narrow confidence interval indicates a very good fit, leading us to believe, that relations with argument distances upwards of $33$ tokens are misclassified by both methods with a significant probability.
If this fact is detrimental to the quality of generated business process models is interesting, but out of scope for this paper.

The \emph{Same Gateway} relation shows frequent misclassification by the \emph{BoostRelEx} method, something that was already evident in Figure~\ref{fig:radar-scenario-4-5-6}.
\emph{BoostRelEx} seems to be very sensitive to the distance between arguments for this relation, more often misclassifying, or outright not recognizing examples, as soon as the distance in tokens exceeds $15$ tokens.
\emph{RuleRelEx} is significantly more robust in this regard, and able to correctly identify \emph{Same Gateway} relations more often than not, until the distance between their arguments exceeds $32$ tokens.
The fit produces very wide confidence intervals for both approaches, something that could be fixed with more examples for this relation, given a larger dataset.

Relations of type \emph{Further Specification} can be extracted by \emph{BoostRelEx} with very high precision and recall.
This is already shown in Figure~\ref{fig:radar-scenario-4-5-6}, where the $F_1$ score for \emph{Further Specification} is given as $0.93$.
The logistic regression fit estimates that there is no correlation between argument distance and correctness. 
Yet, a very wide confidence interval for distances upwards of $10$ tokens leaves open the possibility that there is a correlation given more examples.
While \emph{RuleRelEx} predicts more \emph{Further Specification} relations erroneously than \emph{BoostRelEx}, it is able to classify the majority (distances $0$ -- $6$ tokens) correctly. 
This leads to very similar performance overall, as shown in Figure~\ref{fig:radar-scenario-4-5-6}.

In summary, we expect performance improvements for both  \emph{BoostRelEx} and \emph{RuleRelEx}, if precision and recall for longer distance relations is improved.
Moreover, since our machine learning based RE method outperforms the rule based RE method, in the best case, and is equivalent in the worst case, we can answer \ref{RQ1} with \emph{Yes}. 
Our in-depth evaluation shows, that \emph{BoostRelEx} is fairly robust in dealing with long relations, and only is beaten by \emph{RuleRelEx} on the \emph{Same Gateway} relation, which matters not as much overall, given the small number of examples for this relation.

\section{Conclusion and Future Work}\label{sec:conclusion}
In this paper we extend the task of business process information extraction by ER. 
We enrich PET with entity identity information and propose an extraction approach based on pretrained end-to-end neural coreference resolution.

Motivated by benefits regarding rapid adaption to new data, domains, or tag sets, we propose a novel gradient boost based approach for the relation extraction task. 
We show that our proposed method is able to produce equivalent or better results in the end-to-end setting, and significantly outperform the baseline given higher quality inputs.
We show that PET is not yet extensive enough for training a state-of-the-art deep learning approach from the NLP domain, Jerex, even though this approach achieves state-of-the-art results on other, bigger benchmark datasets of a related task.
Finally, we discuss traits of the PET dataset that are detrimental to prediction quality, e.g., high linguistic variance, and distance between relation arguments.
Our experiments attest to the phenomenon of error propagation, i.e., errors made in early steps are amplified in later ones. 
Thus, we plan to incorporate joint models for extracting mentions, relations, and for resolving process entities, since they are trained to solve these three tasks simultaneously, and mitigate the error propagation effect. 
While Jerex did not produce high quality predictions, it, and similar approaches, are predetermined for application in the task of business process generation from natural language text.
Therefore, further research into applying deep learning in the low data domain of BPM is needed.
We plan to improve performance of the entity resolution module, e.g., by incorporating in-domain knowledge, like mention information from previous steps. 
Additionally, fine-tuning the pretrained neural coreference resolver, by training it on in-domain data is a potential way to improve performance further.
Finally, best practises recommend the use of micro $F_1$ scores for judging the quality of predictions in the business process information extraction task.
While this is certainly a useful metric, we suspect it may not capture the needs of down stream tasks and users entirely. 
We plan to investigate alternative metrics, and their correlation with expectation towards extraction modules by humans.

\bibliographystyle{splncs04}
\bibliography{literature}

\end{document}